\documentclass[10pt,twocolumn,letterpaper]{article}

\usepackage{cvpr}
\usepackage{times}
\usepackage{epsfig}
\usepackage{graphicx}
\usepackage{amsmath}
\usepackage{amssymb}

\usepackage{multirow}
\usepackage{booktabs}
\usepackage{tabularray}
\usepackage{xcolor}
\usepackage{colortbl}
\usepackage{sidecap}
\usepackage[symbol]{footmisc}
\usepackage[accsupp]{axessibility} 
\usepackage[misc]{ifsym}

\usepackage[pagebackref=true,breaklinks=true,letterpaper=true,colorlinks,bookmarks=false]{hyperref}

\begin{document}

\title{FedSIS: Federated Split Learning with Intermediate Representation Sampling for Privacy-preserving Generalized Face Presentation Attack Detection}

\author{Naif Alkhunaizi\textsuperscript{*}, Koushik Srivatsan\textsuperscript{*}, Faris Almalik\textsuperscript{*}, Ibrahim Almakky, Karthik Nandakumar\\
Mohamed Bin Zayed University of Artificial Intelligence\\
Abu Dhabi, United Arab Emirates\\
{\tt\small \{naif.alkhunaizi, koushik.srivatsan, faris.almalik\}@mbzuai.ac.ae} \\ 
\tt\small{\{ibrahim.almakky, karthik.nandakumar\}@mbzuai.ac.ae}}

\maketitle
\thispagestyle{empty}
\footnotetext[1]{ Equal contribution.}
\footnotetext[0]{ Accepted at IJCB 2023.}

\begin{abstract}
   Lack of generalization to unseen domains/attacks is the Achilles heel of most face presentation attack detection (FacePAD) algorithms. Existing attempts to enhance the generalizability of FacePAD solutions assume that data from multiple source domains are available with a single entity to enable centralized training. In practice, data from different source domains may be collected by diverse entities, who are often unable to share their data due to legal and privacy constraints. While collaborative learning paradigms such as federated learning (FL) can overcome this problem, standard FL methods are ill-suited for domain generalization because they struggle to surmount the twin challenges of handling non-iid client data distributions during training and generalizing to unseen domains during inference. In this work, a novel framework called \underline{Fed}erated \underline{S}plit learning with \underline{I}ntermediate representation \underline{S}ampling (FedSIS) is introduced for privacy-preserving domain generalization. In FedSIS, a hybrid Vision Transformer (ViT) architecture is learned using a combination of FL and split learning to achieve robustness against statistical heterogeneity in the client data distributions without any sharing of raw data (thereby preserving privacy). To further improve generalization to unseen domains, a novel feature augmentation strategy called intermediate representation sampling is employed, and discriminative information from intermediate blocks of a ViT is distilled using a shared adapter network. The FedSIS approach has been evaluated on two well-known benchmarks for cross-domain FacePAD to demonstrate that it is possible to achieve state-of-the-art generalization performance without data sharing.\\
   Code: \url{https://github.com/Naiftt/FedSIS}
\end{abstract}

\section{Introduction}

The advent of deep learning has enabled the development of highly accurate face recognition (FR) systems \cite{liu2017sphereface, deng2019arcface, sun2020circle} and made face as the preferred biometric trait in applications ranging from unlocking of personal devices to authentication of financial transactions. However, FR systems are vulnerable to attackers who aim to gain unauthorized access by spoofing the identity of a bonafide individual. Such attacks are called face presentation attacks, where the adversaries use presentation attack instruments (PAIs) such as printed photos, replayed videos, or 3D synthetic masks \cite{yu2022deep} to fool the FR system. Thus, \emph{face anti-spoofing} (FAS) or \emph{face presentation attack detection} (FacePAD) is a critical requirement for securing FR systems.

\begin{figure}
    \centering    \includegraphics[width=0.85\linewidth]{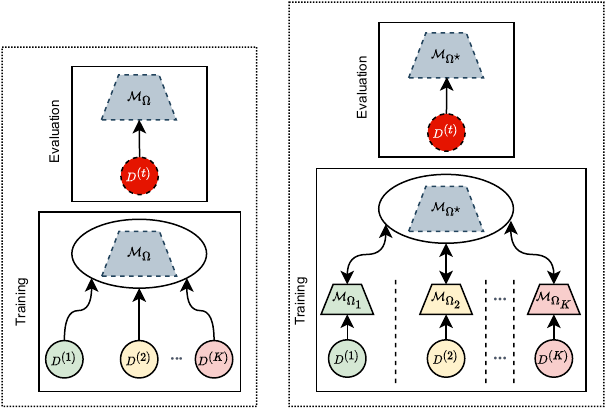}
    \caption{\textbf{Left:} In conventional domain generalization, a single model is trained on $K$ source domains and evaluated on an unseen target domain. \textbf{Right:} In federated domain generalization, each client has a local network trained on the local dataset (source domain), and a global model is constructed by the server (aggregating all the local networks) and evaluated on the unseen domain.}
    \label{fig:intro}
\end{figure}

Many FacePAD algorithms based on convolutional neural network (CNN) \cite{zhang2020face, liu2020disentangling, yu2020face, wang2022disentangled, yu2020searching, yu2020fas, wang2022patchnet} and vision transformer (ViT) \cite{ming2022vitranspad, huang2021multi, wang2022face, wang2022learning} architectures have shown good accuracy in the closed-world (in-domain) setting, where it is assumed that the train and test distributions are the same. However, the closed-world setting is impractical because it is not possible to predict the environmental and capture conditions (e.g., illumination, pose, expression, and camera properties) as well as the PAI that will be encountered by the FacePAD algorithm during deployment. Hence, there is a need for \emph{cross-domain} FacePAD solutions that generalize well to unknown target distributions.

Several methods have been proposed recently for cross-domain FacePAD \cite{shao2019multi, wang2022domain, shao2020regularized, huang2022adaptive, liao2023domain, Sun2023Mar}. These methods can be grouped into three categories based on the problem formulation: unsupervised domain adaptation (UDA) \cite{tzeng2017adversarial, ghifary2016deep, hu2018duplex, li2018unsupervised, wang2020unsupervised, wang2020cross, wang2019improving, jia2021unified, zhou2022generative, yue2022cyclically}, few-shot learning \cite{liu2019deep, qin2020learning, perez2020learning, huang2022adaptive}, and domain generalization (DG) \cite{shao2019multi, shao2020regularized, chen2021generalizable, liu2021dual, wang2021self, liu2021adaptive, liu2022feature, jia2020single, wang2022domain, 10.1007/978-3-031-20065-6_24, liao2023domain}. UDA and few-shot learning methods assume access to the target dataset either in the form of a large set of unlabeled samples (UDA) or a few labeled samples (few-shot). Domain generalization methods propose to learn discriminative domain-invariant features from multiple source domains that generalize to the unseen target domain. Even though the zero-shot DG-FacePAD formulation is the most challenging, it offers a more practical solution for real-world applications. 

The key to the success of DG-FacePAD algorithms is the availability of \emph{large} (to prevent overfitting) and \emph{diverse} (covering different environmental/acquisition conditions and PAIs) datasets for model training. Except for \cite{shao2021federated, shao2022federated}, almost all existing approaches for DG-FacePAD assume that data from multiple source domains is available with a single (central) entity for training (see Fig. \ref{fig:intro}). However, this is an unrealistic assumption given the growing data privacy concerns \cite{gdpr}, which preclude the collection/sharing of large FacePAD datasets. Hence, there is a need for a collaborative training framework where multiple entities having diverse training data can jointly learn a generalizable FacePAD model without sharing sensitive information.  

Though several techniques such as federated learning (FL) \cite{mcmahan2017communication} and split learning (SL) \cite{split_learing} have been proposed in the machine learning community to enable collaborative training by multiple clients without any data sharing, the efficacy of these methods degrade significantly in the heterogeneous/non-iid (the data across clients are not independent and identical distributed) setting. Achieving domain generalization in FL \cite{Liu2021,CCST} is an even more challenging problem because not only are the clients' distributions non-iid during training, but the samples presented during inference are also out-of-domain (target/test distribution is different from any of the client training data distributions). 

In this work, we handle the problem of federated generalized FacePAD (Fed-DG-FacePAD), where each client is assumed to have non-iid bonafide and attack presentation data for training (which cannot be shared for privacy reasons), and the test distribution is unseen. To the best of our knowledge, only the work of Shao et al. \cite{shao2021federated, shao2022federated} has dealt with the same problem setting and proposed a solution based on feature disentanglement and standard federated learning. In contrast, this work advances the state-of-the-art in Fed-DG-FacePAD by making the following contributions:

\begin{itemize}
    \item Inspired by \cite{park2021federated}, we propose a collaborative learning framework for FacePAD based on a hybrid ViT architecture, which consists of a convolution-based tokenizer, self-attention-based feature encoder, and linear classifier. While the tokenizer and classifier components are domain/client-specific and learned in a federated way, the shared feature encoder is learned using split learning and serves to align the source domains.
    
    \item To further enhance the generalizability of the above framework, we propose a novel feature augmentation strategy, where representations learned by intermediate blocks of the ViT are sampled and processed through a shared adapter network before classification. 
    
    \item We empirically demonstrate that the proposed \textbf{FedSIS} approach can outperform even state-of-the-art centralized DG-FacePAD methods while ensuring data privacy for the participating clients. 
\end{itemize}

\section{Related Work}
\subsection{Domain Generalization for FacePAD}
The problem of DG-FacePAD was first considered in \cite{shao2019multi}, where a multi-adversarial discriminative DG framework was proposed to learn a shared feature space from multiple source domains that generalizes to an unseen target domain. In \cite{wang2022domain}, the concept of separating the features into style and content components was introduced and a contrastive learning strategy was applied to emphasize liveness-related style information. A fine-grained meta-learning-based approach was introduced in \cite{shao2020regularized} by simulating the domain shift during training. Recently, ViT-based approaches have shown improved generalization performance. For example, ViTs with ensemble adapter modules and feature-wise transformation layers were employed in \cite{huang2022adaptive} for adapting to the target domain. In \cite{liao2023domain}, two additional losses were proposed to enforce a domain-invariant attack type separation and to push the bonafide representations from multiple domains to be compact. Instead of aiming to learn a domain invariant feature-space, Sun et al. \cite{Sun2023Mar} target domain separability while also aligning the trajectory from live to spoof. They formulate this strategy as an invariant risk minimization problem to learn domain-variant features and domain-invariant classifiers. Although the above methods show promising cross-domain performance, they rely on centrally aggregating the source domain datasets, which leads to privacy concerns.
\subsection{Collaborative Learning}
Federated learning (FL) \cite{FedAvg} is a collaborative learning approach that allows the training of a global model by exchanging model parameters without sharing raw data among clients, thus addressing data privacy concerns. FedAvg \cite{FedAvg} is the most common FL method, which has proven to perform well under iid settings. However, recent works such as \cite{scaffold,Fedprox,harmofl,moon} have focused on non-iid settings. The problem of domain generalization within the FL paradigm was considered in \cite{Liu2021,CCST}.  To better generalize to unseen domains, \cite{Liu2021} focuses on transmitting distribution information across clients while still adhering to data privacy constraints. In \cite{CCST}, cross-client style transfer was proposed to facilitate the integration of all client styles into each client without sharing the local data among the clients, thereby enhancing the ability of local models to accommodate multiple client styles and avoiding model biases. In \cite{shao2021federated, shao2022federated}, both federated FacePAD and Fed-DG-FacePAD are considered. However, the feature disentanglement approach proposed in these works is cumbersome and requires depth map as auxiliary supervision, soft subspace orthogonality constraints to separate domain-specific and domain-invariant features, and autoencoders to ensure reliable reconstruction from all the features.

Split learning (SL) \cite{split_learing} is another collaborative learning approach that aims to ensure privacy by dividing a deep neural network classifier into sub-networks that are spread across different clients. Specifically, the U-shaped SL configuration \cite{U-shaped} consists of three sequential sub-networks across two clients. While the first and last sub-networks are with the data owner, the middle sub-network remains with the other computing client. This configuration ensures that the data owner can collaboratively train the model together with the computing client without sharing raw data or labels. Recently, Park et al. \cite{park2021federated} have demonstrated the effectiveness of combining FL and SL for training ViT models, leveraging their decomposable block structure. 

\section{Methodology}
\begin{figure*}[t]
    \centering    \includegraphics[width=0.85\linewidth]{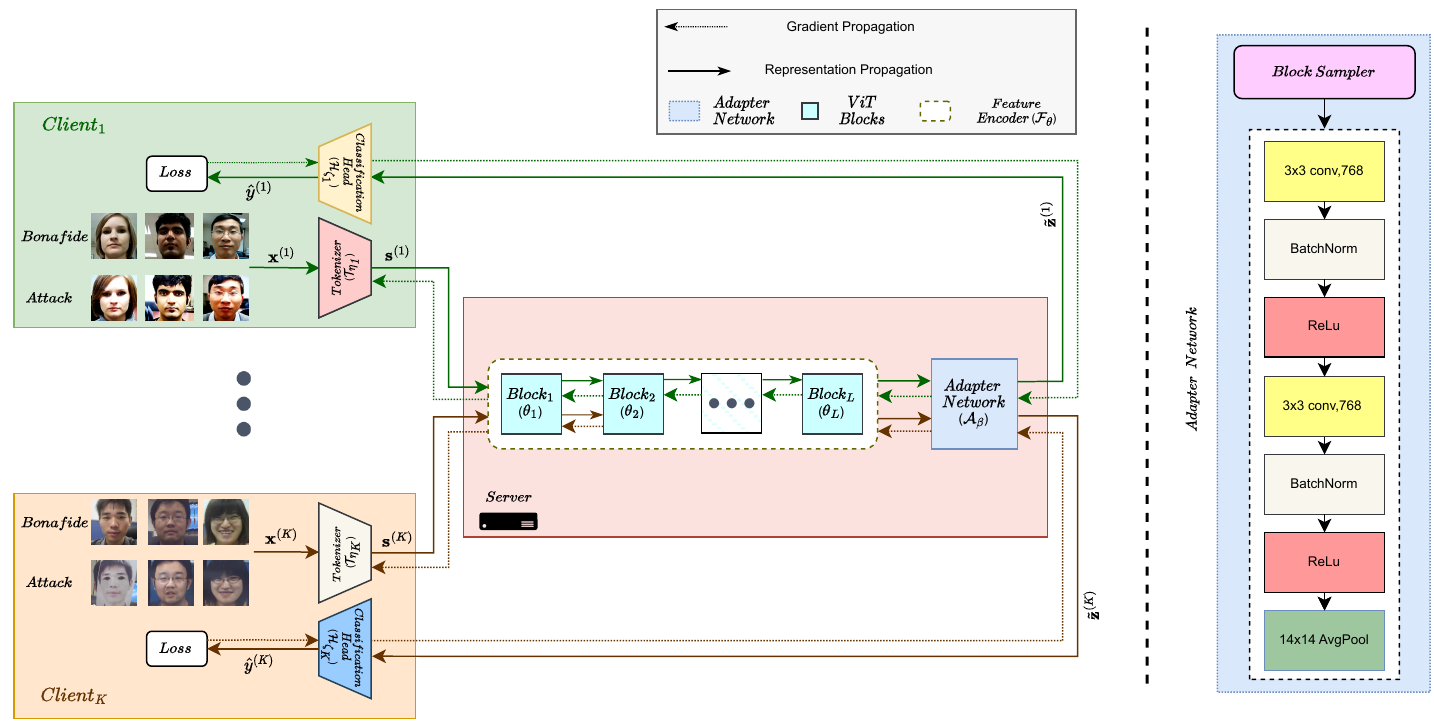}
    \caption{FedSIS framework for privacy-preserving generalized face presentation attack detection is shown on the \textbf{Left}. The clients involved in the training phase have different data distributions and do not share raw image data. Each client encodes the local input images ($\mathbf{x}$) as patch tokens ($\mathbf{s}$) and transmits them to the server. While the ViT-based feature encoder on the server facilitates alignment between the multiple domains, the block sampler and shared adapter network (shown on the \textbf{Right}) act as a feature augmentation scheme to improve generalization. The pseudo-class tokens ($\tilde{\mathbf{z}}$) generated by the adapter are returned to the client for final prediction ($\hat{y}$) using a linear classifier. The shared feature encoder ($\mathcal{F}_{\theta}$) and adapter ($\mathcal{A}_{\beta}$) on the server are trained via split learning and the client-specific tokenizers ($\mathcal{T}_{\eta_{k}}$) and classification heads ($\mathcal{H}_{\zeta_{k}}$) are trained via federated learning.}
    \label{fig:method}
\end{figure*}

\noindent \textbf{Notations}: Let $\mathcal{X} \subseteq \mathbb{R}^{H \times W \times C}$ denote the input face image space, where $H$ and $W$ represent the image height and width, respectively, and $C$ is the number of channels. Let $\mathcal{Y} = \{0,1\}$ be the output label space, where $0$ and $1$ represent \emph{attack} and \emph{bonafide} presentations, respectively. In FacePAD, the goal is to learn a classification model $\mathcal{M}_{\Omega}:\mathcal{X} \rightarrow \mathcal{Y}$ that predicts whether a given face image is a bonafide or attack presentation, where $\mathcal{M}$ denotes the model architecture and $\Omega$ represents the model parameters. Let $Q:P(\mathbf{X},Y)$ denote the joint distribution of random variables $\mathbf{X}$ and $Y$, where $\mathbf{X}\in\mathcal{X}$ and $Y\in\mathcal{Y}$. Let $D^{(*)} = \left\{\left(\mathbf{x}_n^{(*)},y_n^{(*)}\right)\right\}_{n=1}^{N^{(*)}}$ be a set of $N^{(*)}$ samples drawn from the distribution $Q^{(*)}$. Let $f \circ g = g(f(\cdot))$ denote a composition of two functions.
\noindent \textbf{Problem Statement}: Suppose that we are given training samples from $K$ source domains $\mathcal{D}=\{D^{(1)},D^{(2)},\cdots,D^{(K)}\}$, where $D^{(k)}$ is drawn from $Q^{(k)}$ and $Q^{(k)} \neq Q^{(j)} ~ \forall ~ k \neq j, ~k,j = 1,2,\cdots,K$. In DG-FacePAD, the classifier is usually learned by minimizing the following objective function:

\begin{equation}
    \min_{\Omega} \sum_{k=1}^{K} \frac{1}{N^{(k)}} \sum_{n=1}^{N^{(k)}} \mathcal{L}\left(\mathcal{M}_{\Omega}(\mathbf{x}_n^{(k)}),y_n^{(k)}\right),
    \label{eq:DG}
\end{equation}

\noindent where $\mathcal{L}$ is a loss function between the model predictions and the ground truth labels. Furthermore, the above classifier is evaluated based on test samples from an unseen domain (thus requiring \emph{zero-shot} generalization), and the test accuracy is defined as $\mathbb{E}_{(\mathbf{x},y) \sim Q^{(t)}} I(\mathcal{M}_{\Omega}(\mathbf{x})=y)$. Here, the test distribution $Q^{(t)}$ is different from all the $K$ source distributions, i.e.,  $Q^{(t)} \neq Q^{(k)},\forall ~ k = 1, 2,\cdots,K$ and $I$ denotes an indicator function ($1$ if the condition is true and $0$ otherwise). In Fed-DG-FacePAD, we further assume that the training data from the $K$ source domains are available with $K$ different entities (referred to as \emph{clients} in this work) and cannot be pooled centrally due to privacy constraints. Hence, the objective in eq. \ref{eq:DG} cannot be optimized directly.

\subsection{Vision Transformer (ViT)}

We choose a ViT-based architecture as the core for the FacePAD task. A ViT-based classification model ($\mathcal{M}$) typically consists of three components: tokenizer, feature encoder, and classification head. The tokenizer ($\mathcal{T}_{\eta}:\mathcal{X}\rightarrow \mathcal{S}, \mathcal{S} \subseteq \mathbb{R}^{S \times d}$) divides an image into $S$ non-overlapping patches and represents each patch using a $d$ dimensional feature representation called \textit{patch token}. In a vanilla ViT, the raw image patches are flattened and passed through a trainable linear projection layer and appended with a  learnable positional encoding vector (to preserve positional information) to generate the patch tokens. The feature encoder component ($\mathcal{F}_{\theta}:\mathcal{S}\rightarrow \mathcal{Z}, \mathcal{Z} \subseteq \mathbb{R}^{d}$) produces a global representation of the input image called ``classification token'' (\textit{cls} token) by propagating the patch tokens through a sequence of $L$ ViT blocks (self-attention modules) and distilling the knowledge into the \textit{cls} token. The classification head ($\mathcal{H}_{\zeta}:\mathcal{Z}\rightarrow \mathcal{Y}$) is typically a Multilayer Perceptron (MLP) trained using standard loss functions such as cross-entropy loss. Thus, the ViT FacePAD classifier can be summarized as $\mathcal{M}_{\Omega} = (\mathcal{T}_{\eta} \circ \mathcal{F}_{\theta} \circ {H}_{\zeta})$, where $\Omega = [\eta,\theta,\zeta]$.

\subsection{Federated Learning (FL)}

As explained in the problem statement, the data required to train a generalized FacePAD classifier is assumed to come from $K$ source domains, which are available with $K$ different clients. Since this data cannot be pooled centrally to train the model, we employ FL \cite{mcmahan2017communication} to learn the model in a distributed way. Typically, FL training consists of local model update and model aggregation. In a local update, each client individually optimizes the following objective:

\begin{equation}
    \min_{\Omega_k} \frac{1}{N^{(k)}} \sum_{n=1}^{N^{(k)}} \mathcal{L}\left(\mathcal{M}_{\Omega_k}(\mathbf{x}_n^{(k)}),y_n^{(k)}\right).
    \label{eq:FL}
\end{equation}

\noindent After a fixed number of local updates, all the clients send the locally updated model parameters to a server, which aggregates the local model parameters $\Omega_k$ in each \textit{unifying round} ($r_{uni}$) to obtain the global model parameters as:

\begin{equation}
    \Omega = \sum_{k=1}^{K} \rho_k \Omega_k,
    \label{eq:FL-aggregation}
\end{equation}

\noindent where $\rho_k$ is the weight assigned to the client $k$. The global parameters $\Omega$ are broadcast to all clients participating in the protocol at the beginning of the following round. This process is repeated until the global model converges to an acceptable level of performance. The most commonly used algorithm for aggregation is called federated averaging (FedAvg), where $\rho_k = N^{(k)}/N$ and $N = \sum_{k=1}^K N^{(k)}$.

\subsection{Proposed FedSIS Framework}
\label{sec:FedSIS_method}

Since the goal is to achieve federated domain generalization for FacePAD, both the core ViT architecture and the FL training strategy need to be modified suitably for this task. Hence, we propose a new framework called \textbf{FedSIS}, which employs a modified ViT architecture and a federated split learning algorithm for training. 

\noindent \textbf{Hybrid ViT (ViT-H)}: The vanilla ViT architecture has worked well in many image recognition applications (including FacePAD) primarily due to its ability to capture long-range dependencies among different image patches, thus encoding more global characteristics of the image. However, good FacePAD performance also requires accurate characterization of micro-texture details \cite{wang2022face}. Hence, a CNN-based tokenizer $\mathcal{T}$ \cite{dosovitskiy2020image} is more appropriate for the FacePAD task. The same feature encoder $\mathcal{F}$ used in vanilla ViT can be retained. Finally, since we are dealing with a binary classification task, it suffices to use a linear classifier as the classification head ($\mathcal{H}$), which is applied to the cls token. We refer to the above architecture as ViT-H.

\noindent \textbf{ViT-H with Intermediate Representation Sampling (ViT-H-IS)}: Typically, ViTs rely solely on the cls token output by the last block while ignoring the patch and cls tokens produced by the intermediate ViT blocks. However, it has been argued recently that the texture information is better encapsulated in the patch tokens generated by the intermediate ViT blocks \cite{faris_paper,rouqiah_paper}. The challenge is that it is not possible to determine which intermediate block produces the best representation for the FacePAD task,
because it appears to be data-dependent (see supplementary material). To overcome this problem, we propose the following solution. 

Suppose that for a given image $\mathbf{x} \in \mathcal{X}$, the tokenizer $\mathcal{T}_{\eta}$ outputs the patch tokens represented as $\mathbf{s} \in \mathcal{S}$. Note that the feature encoder $\mathcal{F}_{\theta}$ can be considered as a sequential composition of $L$ self-attention ViT blocks, i.e., $\mathcal{F}_{\theta} = \mathcal{F}_{\theta_1} \circ \mathcal{F}_{\theta_2} \circ \cdots \mathcal{F}_{\theta_L}$. Instead of passing the input patch tokens through the entire feature encoder $\mathcal{F}_{\theta}$ to obtain the final cls token $\mathbf{z}$, we employ a \textit{block sampler} to randomly choose an intermediate self-attention block $\ell \in [1,2,\cdots,L]$ for each training round. We propagate the input patch tokens $\mathbf{s}$ only through the first $\ell$ self-attention blocks to obtain intermediate patch tokens $\mathbf{s}_{\ell} = \mathcal{F}_{\theta_{[1:\ell]}} = \mathcal{F}_{\theta_1} \circ \mathcal{F}_{\theta_2} \circ \cdots \mathcal{F}_{\theta_{\ell}}(\mathbf{s})$. In ViT-H-IS, this sampled intermediate representation $\mathbf{s}_{\ell}$ is processed through an adapter network $\mathcal{A}_\beta:\mathcal{S} \rightarrow \mathcal{Z}$ to obtain a pseudo-class token $\tilde{\mathbf{z}} = \mathcal{A}_\beta(\mathbf{s}_{\ell})$. Note that a single adapter network $\mathcal{A}_\beta$ is shared by all the ViT blocks. The adapter network serves as an additional distillation path to capture discriminative information required for FacePAD from the intermediate patch tokens.

While the classification head $\mathcal{H}_{\zeta}$ receives the cls token $\mathbf{z}$ from the last self-attention block in ViT and ViT-H, it receives the pseudo-class token $\tilde{\mathbf{z}}$ in ViT-H-IS. The proposed intermediate representation sampling approach effectively utilizes intermediate features in a ViT feature encoder, which are completely ignored in ViT or ViT-H, but may be critical for the FacePAD task. Furthermore, sampling these intermediate representations from different blocks, rather than relying solely on an individual block’s features or the final cls token, serves as an effective \emph{feature augmentation} strategy, enhancing the domain generalization capability. 

\begin{table*}[!t]
\centering \small
\caption{Cross-domain performance of centralized and federated learning methods on Benchmark 1 using CelebA-Spoof as an auxiliary dataset. Each experiment was repeated 3 times with different seeds and mean values are reported.}
\label{tab:cross_domain_mcio}
\setlength{\tabcolsep}{3pt}
      \scalebox{0.78}[0.80]{
	\begin{tabular}{llccccccccccccccccc}
	\multicolumn{17}{c}{} \\ 
        \toprule
	  & \multirow{3}{*}{\bf ~~Method~} & \multicolumn{3}{c}{\textbf{OCI} $\rightarrow$ \textbf{M}}  && \multicolumn{3}{c}{\textbf{OMI} $\rightarrow$ \textbf{C}}  &&  \multicolumn{3}{c}{\textbf{OCM} $\rightarrow$ \textbf{I}} && \multicolumn{3}{c}{\textbf{ICM} $\rightarrow$ \textbf{O}} && \textbf{Avg.} \\
	 \cmidrule{3-5} \cmidrule{7-9} \cmidrule{11-13} \cmidrule{15-17} 
	 && \multirow{2}{*}{~HTER~} & \multirow{2}{*}{~AUC~} & TPR@ && \multirow{2}{*}{~HTER~} & \multirow{2}{*}{~AUC~} & TPR@ && \multirow{2}{*}{~HTER~} & \multirow{2}{*}{~AUC~} & TPR@ && \multirow{2}{*}{~HTER~} & \multirow{2}{*}{~AUC~} & TPR@ && \multirow{2}{*}{~HTER~} \\
	 &&&& ~FPR=$1\%$~ &&&& ~FPR=$1\%$~ &&&& ~FPR=$1\%$~ &&&& ~FPR=$1\%$~ \\ 
    \midrule
     & ~~ViT (ECCV' 22) \cite{huang2022adaptive}             & {1.58} & {99.68} & {96.67} && 5.70 & 98.91 & 88.57 && 9.25 & 97.15 & 51.54 && 7.47 & {98.42} & {69.30} && 6.00 \\      
     & ~~ViT-H        & 7.81 & 97.76  & 78.33  && 2.64  & 99.34  & 88.57  & & 16.08 & 90.13  & 4.10 & & {7.13}  & 97.84 & 59.76 & & 8.41 \\
     \multirow{-3}{*}{Centralized}
     & ~~ViT-H-IS       & 2.70 & 99.30  & 83.10  &&  2.90 & 99.58  & 89.71  && 5.44 &  99.19 & 74.07 && 11.19  & 95.70  & 40.60 && 5.56  \\
     \midrule
     & ~~FedAVG \cite{FedAvg}         & $8.69 $ & $97.80 $ & $77.77 $ && $1.67 $ & $99.50 $ & $95.27 $ && $20.96 $ & $84.08 $ & $3.11 $  && $\textbf{9.30 }$  &  $95.60 $ & 28.03  & & $10.15$  \\
     
     & ~~FedProx \cite{Fedprox}       & $7.64 $ & $97.61 $ & $71.11 $ && $1.98 $  & $99.75 $ &  $97.38 $ && $24.55 $ & $81.31 $  & $1.28$ & & $17.04$ & $91.58 $ & $31.55 $ & & $12.80$  \\
     & ~~MOON \cite{moon} & $7.91$  & $97.99$  & $75.0$ & & $1.98$ & $99.818 $ & $94.28 $ & & $21.49 $  & $85.19 $ & $10.76 $ && $11.12 $  & $95.10 $ & $20.30 $ & & $10.62 $  \\
      & ~~FeSTA \cite{park2021federated} & $7.81$  & $97.00$  & $53.33 $ & & $7.82 $ & $97.49 $ & $65.24 $ & & $9.10 $  & $96.15 $ & $40.26 $ && $17.84 $  & $89.74 $ & $15.30 $ & & $10.64 $  \\
    \rowcolor{yellow!15}
     \multirow{-5}{*}{Federated} 
     & ~~FedSIS (Ours)        & $\textbf{1.22} $ & ${99.66}$ & ${96.11}$ & & $\textbf{1.39}$ & ${99.67}$ & ${96.66} $ & & $\textbf{3.36}$ & ${99.18} $ & ${75.12} $ && $9.52$ & ${96.25}$ & ${40.23}$  & & $\textbf{3.91}$ \\
    \bottomrule
	\end{tabular}
}
\end{table*}


\noindent \textbf{Federated Split Learning}: We now turn our attention to the training strategy. Recent work has suggested that the core challenge in DG-FacePAD is aligning the bonafide-to-attack transition across all domains \cite{Sun2023Mar}. Based on this intuition, we adapt the federated split task-agnostic (FeSTA) learning framework proposed in \cite{park2021federated} to achieve better alignment between domains. While FeSTA was originally proposed for task-agnostic FL, we show that it can be leveraged to achieve domain alignment in DG. 

Federated split learning of the proposed ViT model works as follows. Each client has an individual (client-specific) tokenizer ($\mathcal{T}_{\eta_{k}}$) and classification head ($\mathcal{H}_{\zeta_{k}}$). A shared feature encoder ($\mathcal{F}_{\theta}$) and an adapter network ($\mathcal{A}_{\beta}$) are held by the server and are trained centrally. The server initializes the full model (including tokenizer, feature encoder, adapter network, and classification head) either randomly or via pre-training on a public dataset. It broadcasts the tokenizer and classification head to all the clients to initialize the client-specific networks.

During the training phase, in each collaboration round, a client $k$ samples a training batch $D^{(k)}_B = \left\{\left(\mathbf{x}_n^{(k)},y_n^{(k)}\right)\right\}_{n=1}^{N^{(k)}_B} \subseteq D^{(k)}$, where $N^{(k)}_B$ is the batch size. The input images in the batch are tokenized using $\mathcal{T}_{\eta_{k}}$ to obtain the patch token representations $\left\{\mathbf{s}_n^{(k)}\right\}_{n=1}^{N^{(k)}_B}$, where $\mathbf{s}_n^{(k)} = \mathcal{T}_{\eta_{k}}(\mathbf{x}_n^{(k)}), \forall~n = 1,2,\cdots,N^{(k)}_B$. These patch token representations are then sent to the server.

The block sampler on the server randomly chooses a self-attention block $\ell \in [1,2,\cdots,L]$ for that client and round. It propagates the patch tokens through the first $\ell$ blocks and the adapter network to obtain the pseudo-class tokens $\left\{\tilde{\mathbf{z}}_n^{(k)}\right\}_{n=1}^{N^{(k)}_B}$, where $\tilde{\mathbf{z}}_n^{(k)} =  \mathcal{A}_\beta(\mathcal{F}_{\theta_{[1:\ell]}}(\mathbf{s}_n^{(k)}))$. These pesudo-class tokens are returned to the client $k$, which applies its classification head $\mathcal{H}_{\zeta_{k}}$ to predict the class labels $\hat{y}_n^{(k)} = \mathcal{H}_{\zeta_{k}}(\tilde{\mathbf{z}}_n^{(k)})$ and compute the sample loss $\mathcal{L}(\hat{y}_n^{(k)},y_n^{(k)}$). This completes the forward pass and the learning objective in each round can be summarized as:
\begin{equation}
\begin{split}
        \min_{\eta_k,\theta_{[1:\ell]},\beta,\zeta_k} &\quad \frac{1}{N^{(k)}_B} \sum_{n=1}^{N^{(k)}_B} \mathcal{L}\left(\hat{y}_n^{(k)},y_n^{(k)}\right),\\
         \text{where} &\quad  \hat{y}_n^{(k)} = \mathcal{T}_{\eta_{k}} \circ \mathcal{F}_{\theta_{[1:\ell]}} \circ \mathcal{A}_\beta \circ \mathcal{H}_{\zeta_{k}}(\mathbf{x}_n^{(k)}).
    \label{eq:FedSIS}
\end{split}
\end{equation}
To solve the above optimization problem, the client $k$ computes the gradients and backpropagates them in the reverse order: through the client's classification head, server's adapter network, server's first $\ell$ feature encoder blocks, and the client's tokenizer. Within each collaboration round, the above process is repeated for all the $K$ clients. While the parameters of the classification head, adapter network, and tokenizer are updated after every backward pass, the parameters of the feature encoder are updated only at the end of a collaboration round after aggregating updates from all the $K$ clients. Finally, FedAvg is used to aggregate the parameters of the clients' tokenizers and classification heads after every unifying round ($r_{uni}$), and the server broadcasts the aggregated parameters back to all clients for the following round. Thus, at the end of the training phase, we obtain a single (global) hybrid ViT model together with an adapter network. Note that this global model can be used for inference on out-of-domain data. 

\begin{figure*}[t!]
    \centering{
    \includegraphics[width=0.87\textwidth]{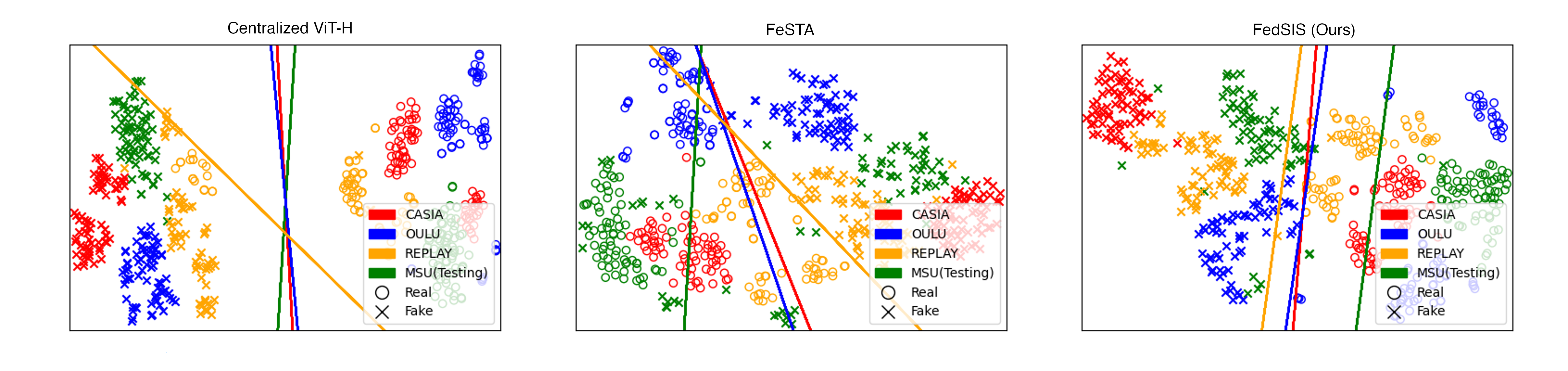}
    }
    \vspace{-1.0em}
    \caption{Visualization of decision boundaries learned by a linear SVM classifier after dimensionality reduction (of the input presented to the classification head) using t-SNE. These results are based on representative sub-samples from the MCIO datasets (Benchmark 1). In the absence of a shared feature encoder, the decision boundaries of different domains are not aligned (Left), confirming the hypothesis propounded in \cite{Sun2023Mar}. A shared feature encoder trained using split learning leads to better domain alignment as seen in the case of FeSTA (Middle). FedSIS (Right) further aligns the domain decision boundaries and makes them more robust to distribution shifts.} 
    \label{fig:decision_boundary}
\end{figure*}

Some key points about the FedSIS framework need to be highlighted. Firstly, the predictions made by FedSIS are non-deterministic because intermediate representation sampling is also applied during inference. Secondly, the shared server network in FedSIS serves the purpose of aligning the bonafide to attack presentation trajectories in the feature space. This is especially critical because we use a simple linear classifier for the final classification decision. If the decision boundaries of different domains are not aligned, FedAvg of the domain-specific classification heads is unlikely to work. Finally, as shown in Fig. \ref{fig:decision_boundary}, both FeSTA and FedSIS can facilitate better domain alignment in the feature space. While FeSTA utilizes only the last cls token, the proposed FedSIS framework leverages the intermediate patch tokens that are sampled from different depths. This not only encapsulates critical texture information but also challenges the classification heads to learn more generalized decision boundaries. Thus, the FedSIS framework facilitates more robust domain generalization for the FacePAD task without compromising on data privacy.

\section{Experiments}
\subsection{Datasets}
We evaluate the proposed framework on 2 well-known benchmarks (consisting of 7 datasets in total) following the leave-one-domain-out evaluation protocols in \cite{shao2019multi, jia2020single, wang2022domain, huang2022adaptive}. In \textbf{Benchmark 1}, we evaluate on MSU-MFSD \textbf{(M)} \cite{7031384}, CASIA-MFSD \textbf{(C)} \cite{6199754}, Idiap Replay Attack \textbf{(I)} \cite{6313548}, and OULU-NPU \textbf{(O)} \cite{7961798} datasets. Specifically, \textbf{OCI} $\rightarrow$ \textbf{M} represents the scenario where \textbf{O}, \textbf{C}, and \textbf{I} datasets are considered as individual clients (source domains) and  \textbf{M} is the target client (unseen domain). In \textbf{Benchmark 2}, we evaluate our method on WMCA \textbf{(W)} \cite{8714076}, CASIA-CeFA \textbf{(C)} \cite{9423056, liu2021cross}, and CASIA-SURF \textbf{(S)} \cite{8995504, zhang2019dataset}, which are large-scale FacePAD datasets involving a larger number of samples and identities compared to the datasets in \textbf{Benchmark 1}. Similar to \cite{huang2022adaptive}, we employ CelebA-Spoof \cite{zhang2020celeba} as an auxiliary dataset for both benchmarks. However, to ensure a fair comparison with some reported results, we also consider the case where CelebA-Spoof is not used.

\begin{table*}[!t]
\centering \small
\caption{Effect of excluding CelebA-Spoof dataset on cross-domain performance in Benchmark 1. First 6 rows represent recent centralized DG-FacePAD methods reported in the literature and the next 2 rows are FL methods. FedSIS$^{*}$ indicates not using CelebA-Spoof both at the pre-training and collaborative training stages. For FedSIS$^{*}$, we run the experiment 3 times and report the mean performance.
}
\label{tab:effect_of_celeba_mcio}
\setlength{\tabcolsep}{3pt}
      \scalebox{0.8}[0.80]{
	\begin{tabular}{llccccccccccccccccc}
	\multicolumn{16}{c}{} \\ 
        \toprule
	  & \multirow{3}{*}{\bf ~~Method~} & \multicolumn{3}{c}{\textbf{OCI} $\rightarrow$ \textbf{M}}  && \multicolumn{3}{c}{\textbf{OMI} $\rightarrow$ \textbf{C}}  &&  \multicolumn{3}{c}{\textbf{OCM} $\rightarrow$ \textbf{I}} && \multicolumn{3}{c}{\textbf{ICM} $\rightarrow$ \textbf{O}} && \textbf{Avg.} \\
	 \cmidrule{3-5} \cmidrule{7-9} \cmidrule{11-13} \cmidrule{15-17} 
	 && \multirow{2}{*}{~HTER~} & \multirow{2}{*}{~AUC~} & TPR@ && \multirow{2}{*}{~HTER~} & \multirow{2}{*}{~AUC~} & TPR@ && \multirow{2}{*}{~HTER~} & \multirow{2}{*}{~AUC~} & TPR@ && \multirow{2}{*}{~HTER~} & \multirow{2}{*}{~AUC~} & TPR@ && \multirow{2}{*}{~HTER~} \\
	 &&&& ~FPR=$1\%$~ &&&& ~FPR=$1\%$~ &&&& ~FPR=$1\%$~ &&&& ~FPR=$1\%$~ \\ 
    \midrule
    
     &~~ViT (ECCV' 22) \cite{huang2022adaptive}        & 4.75 & 98.79 & 68.33 && 15.70 & 92.76 & 36.43 && 17.68 & 86.66 & 50.77 && 16.46 & 90.37 & 24.23 && 13.64   \\
     &~~SSAN-R (CVPR' 22) ~\cite{wang2022domain}       & 6.67  & 98.75 & -- && 10.00 & 96.67 & -- && 8.88  & 96.79 & -- && 13.72 & 93.63 & -- && 9.80\\
     &~~PatchNet (CVPR' 22) ~\cite{wang2022patchnet}       &  7.10 & 98.46 & -- && 11.33 & 94.58 & -- && 13.40 & 95.67 & -- && 11.82 & 95.07 & -- && 10.90\\
     &~~GDA (ECCV' 22) ~\cite{zhou2022generative}       & 9.20  & 98.00 & -- && 12.20 & 93.00 & -- && 10.00 & 96.00 & -- && 14.40 & 92.60 & -- && 11.45\\
     &~~SA-FAS (CVPR' 23) \cite{Sun2023Mar}       & 5.95 & 96.55 & - && 8.78 & 95.37 & - && \textbf{6.58} & 97.54 & - && 10.00 & 96.23 & - && 7.81 \\
     &~~IADG  (CVPR' 23) \cite{zhou2023instance}     & 5.41 & 98.19 &  - && 8.70 & 96.44 & - && 10.62 & 94.50 & - && \textbf{8.86} & 97.14 & - && 8.39 \\
     \midrule
     & ~~FedPAD (TNNLS' 22) \cite{shao2022federated} & 19.45 & 90.24  & - && 42.27 & 70.49 & - && 32.53  & 73.58 & - &&  34.44 & 71.74 & - && 32.17   \\
     & ~~FedGPAD (TNNLS' 22) \cite{shao2022federated} & 12.73 & 91.25  & - && 28.69 & 80.58 & - && 10.97 & 95.34 & - && 21.95 & 89.85 & -  && 18.59  \\
     \midrule
    \rowcolor{yellow!15}
     &~~FedSIS$^{*}$  (Ours)        & \textbf{2.25} & 99.42 & 72.16 && \textbf{5.75} & 98.49 & 70.77 && 7.39 & 97.44 & 52.52 && 10.68 & 95.80 & 45.57 && \textbf{6.43}  \\
    \bottomrule
	\end{tabular}
}
\end{table*}

\begin{table*}[!t]
\centering \small
\caption{Cross-domain performance of centralized and federated learning methods on Benchmark 2 using CelebA-Spoof as an auxiliary dataset. Each experiment was repeated 3 times with different seeds and mean values are reported.}
\label{tab:cross_domain_wcs}
\setlength{\tabcolsep}{3pt}
      \scalebox{0.85}[0.85]{
	\begin{tabular}{llccccccccccccc}
	\multicolumn{13}{c}{} \\ 
        \toprule
	  & \multirow{3}{*}{\bf ~~Method~} & \multicolumn{3}{c}{\textbf{CS} $\rightarrow$ \textbf{W}}  && \multicolumn{3}{c}{\textbf{SW} $\rightarrow$ \textbf{C}}  &&  \multicolumn{3}{c}{\textbf{WC} $\rightarrow$ \textbf{S}} & & \textbf{Avg.} \\
	 \cmidrule{3-5} \cmidrule{7-9} \cmidrule{11-13} 
	 & & \multirow{2}{*}{~HTER~} & \multirow{2}{*}{~AUC~} & TPR@ & & \multirow{2}{*}{~HTER~} & \multirow{2}{*}{~AUC~} & TPR@ && \multirow{2}{*}{~HTER~} & \multirow{2}{*}{~AUC~} & TPR@ && \multirow{2}{*}{~HTER~} \\
	 &&&& ~FPR=$1\%$~ &&&& ~FPR=$1\%$~ &&&&  ~FPR=$1\%$~ \\ 
    \midrule
     & ~~ViT (ECCV22) \cite{huang2022adaptive}             & 7.98 & 97.97 & 73.61 &&  11.13 & 95.46 & 47.59 && 13.35 &94.13  &  49.97 &&  10.82\\				
     \multirow{-2}{*}{Centralized}
     & ~~ViT-H         & 8.31 &  97.71 & 64.22  &&  15.68 &  92.99 &  48.05  && 16.89 &  89.97 & 32.43 &&   13.63 \\						
     \midrule
     & ~~FedAVG \cite{FedAvg}         & 20.18 & 89.47 & 45.15 && 25.94 & 82.05 & 22.18 && 13.52 & 93.70 & 43.06  && 19.88\\
     & ~~FedProx \cite{Fedprox}       &18.01  & 90.81&54.11  &&26.98 &  81.09& 25.95 && 12.02 & 94.49  &  52.04 && 14.25 \\	
     & ~~FeSTA \cite{park2021federated} &  8.49 &  96.79 & 61.31 && 19.15 & 89.81 & 32.02 && \textbf{9.60}  & 96.80 &61.76  && 12.41\\

     \rowcolor{yellow!15}
     \multirow{-4}{*}{Federated}
     & ~~FedSIS (Ours)        & \textbf{5.84} &98.81 &77.74 && \textbf{14.50} & 93.80 &39.89  && 11.03 & 95.65 &55.797  &&  \textbf{10.45}\\
    \bottomrule
	\end{tabular}
}
\end{table*}

\subsection{Implementation Details}

\noindent\textbf{Model Architecture}: We choose ViT-B/16 implementation in \cite{timm} as the base feature encoder $\mathcal{F}$. This model has $L = 12$ self-attention blocks and $12$ attention heads pretrained on ImageNet. The vanilla ViT model divides a $224 \times 224 \times 3$ RGB image into $16 \times 16$ non-overlapping patches, resulting in $S = 196$ patch tokens that are represented as $d=768$-dimensional vectors. In ViT-H, the tokenizer $\mathcal{T}$ is a truncated ResNet-50 model pre-trained on ImageNet, which is the same as the Hybrid-ViT implementation in \cite{timm}. The number and dimensionality of \textit{patch} tokens in ViT-H are the same as in vanilla ViT. For the shared adapter network $\mathcal{A}$ (shown in Fig. \ref{fig:method} Right), we design a CNN with two convolutional layers, each followed by BatchNorm and ReLU and finally an AvgPool layer. The classification head $\mathcal{H}$ consists of a single linear layer that maps the pseudo class token to the two output classes.

\noindent\textbf{Pre-training}: When CelebA-Spoof dataset is not used, we initialize the tokenizer and feature encoder using ImageNet pre-trained weights and randomly initialize the adapter network and classification head. However, when CelebA-Spoof is used as the auxiliary dataset, we additionally pre-train the full ViT-H-IS model centrally using $90\%$ of the CelebA-Spoof samples and initialize all the experiments with these pre-trained weights. This is motivated by observations in \cite{chen2023importance} on how pre-training in FL can help bridge the performance gap to centralized learning.

\noindent\textbf{Collaborative Training}: During collaborative training, we set the number of clients (source domains) to $K = 3$ for Benchmark 1 and $K = 2$ for Benchmark 2. When CelebA-Spoof is used as the auxiliary dataset, we add an additional client to both benchmarks to utilize the remaining $10\%$ of the CelebA-Spoof samples, thereby increasing the diversity of the training data. Furthermore, we train for $200$ rounds, where the aggregation of tokenizer and classification head parameters happens every $r_{uni} = 10$ rounds via FedAvg. We use Adam optimizer \cite{Adamopt} with weight decay of $10^{-6}$ for both benchmarks and a learning rate of $10^{-6}$ and $10^{-5}$ for Benchmark 1 and Benchmark 2, respectively. We use the standard cross-entropy loss as the training objective. For Benchmark 1, we set a batch size of $6$ for all clients except CelebA-Spoof. Due to the larger number of samples in CelebA-Spoof, we increased its batch size to $32$. This improved overall efficiency and ensured that each client had a similar number of iterations during training, thereby leading to better training stability. For Benchmark 2, we set a batch size of $32$ for all clients. The code was implemented in PyTorch $1.10$, and all experiments were executed on Nvidia A100 GPU with 40 GB memory. 

\subsection{Evaluation Metrics}
As outlined in \cite{huang2022adaptive}, we evaluate the model on three different performance metrics: Half Total Error Rate (HTER), Area Under the Receiver Operating Characteristic (AUC), and True Positive Rate (TPR) at a fixed False Positive Rate (FPR). Note that FPR is nothing but a Bonafide Presentation Classification Error Rate (BPCER), and TPR is the same as (100 - APCER), where APCER is the Attack Presentation Classification Error Rate \cite{standard2017information}. To mitigate the impact of statistical variations and the non-deterministic nature of predictions in FedSIS, each experiment was conducted three times with different seeds, and the mean scores along with their standard deviation (see supplementary) are reported.

\section{Results and Discussion}
\noindent \textbf{Baselines}: Within the FacePAD literature, the closest baseline for the proposed FedSIS framework is FedGPAD \cite{shao2022federated}, which works under the exact same setting. However, to better illustrate the benefits of the FedSIS framework, we also compare it against the following sets of baselines: (i) ViT, ViT-H, and ViT-H-IS trained centrally (pooling all the source domains), and (ii) collaborative learning algorithms like FedAvg \cite{FedAvg}, FedProx \cite{Fedprox}, MOON \cite{moon}, and FeSTA \cite{park2021federated} using the ViT-H architecture as the base model.

\noindent \textbf{Cross-Domain Performance on Benchmark 1}: Table \ref{tab:cross_domain_mcio} summarizes the cross-domain performance of FedSIS against other federated and centralized learning methods on Benchmark 1 when CelebA-Spoof is used as the auxiliary dataset. The results in Table \ref{tab:cross_domain_mcio} indicate that FedSIS outperforms all the federated and centralized learning methods in three out of four target domains. Only for the ICM $\rightarrow$ O scenario, we note that FedSIS has a lower HTER compared to FedAvg (by a small margin of -0.22) and the centralized ViT-H (by a margin of -2.39). We attribute this to the size of the target domain O, which contains three times as many samples as the other three source domains combined. We also make the following observations based on Table \ref{tab:cross_domain_mcio}.

\begin{itemize}
    \item The superior performance of the centrally trained ViT-H-IS model compared to the centralized ViT-H model demonstrates the utility of our intermediate representation sampling approach. The block sampler and shared adapter network used in ViT-H-IS enables the implementation of an effective feature augmentation strategy, which leads to better domain generalization.
    \item The lower average HTER achieved by FedSIS in comparison to centralized ViT-H-IS shows that the performance gain is not merely due to the choice of model architecture but federated split learning also plays a crucial role in boosting the performance. It is also contrary to the conventional wisdom that centralized training serves as an upper bound for FL methods and there is always a privacy-utility trade-off in FL. In this work, we have shown that well-designed collaborative training can effectively overcome this trade-off and achieve good performance while ensuring privacy.
    \item The fact that FedSIS outperforms FL methods like FedProx and MOON proves that accounting for non-iid data distributions during training is insufficient for out-of-domain generalization during inference.
    \item Finally, FedSIS also performs better than FeSTA, which applies the same federated split learning approach based on the ViT-H model. While FeSTA increases the generalization error compared to centrally trained ViT-H, FedSIS decreases the error compared to centralized ViT-H-IS. This highlights the need for co-designing the model architecture and the collaborative learning method to leverage their mutual strengths. 
\end{itemize}

\noindent \textbf{Effect of excluding CelebA-Spoof}: In Table \ref{tab:effect_of_celeba_mcio}, we study the effect of excluding the auxiliary CelebA-Spoof dataset and present a direct comparison to other DG-FacePAD methods from the literature. While FedSIS in Table \ref{tab:cross_domain_mcio} indicates using CelebA-Spoof both at the pre-training and collaborative training stages, FedSIS$^{*}$ in Table \ref{tab:effect_of_celeba_mcio} indicates not using CelebA-Spoof at both these stages. It is interesting to observe that FedSIS$^{*}$ outperforms all the state-of-the-art centralized DG-FacePAD methods by a comfortable margin. Furthermore, FedSIS outperforms FedGPAD by a substantially large margin, which again emphasizes the importance of selecting a suitable model architecture and collaborative training algorithm that complement each other. 

\noindent \textbf{Cross-Domain Performance on Benchmark 2}: To gain further insight into the performance of FedSIS when training with large-scale anti-spoofing datasets, we follow the same leave-one-domain-out strategy and report the results based on the larger W, C, and S datasets in Table \ref{tab:cross_domain_wcs}. While the overall trends are broadly similar to Benchmark 1, we observe that the performance gains are marginal in Benchmark 2. This is to be expected because collaborative learning is more beneficial when each client has limited data. In contrast, when all the clients have access to large datasets locally, there is little to learn from each other.

The supplementary material includes additional ablation studies on: (i) the need for randomly sampling an intermediate block instead of choosing a pre-determined intermediate block, (ii) the impact of restricting block sampling to a subset of blocks, (iii) the optimal choice of the number of unifying rounds, and (iv) the impact of partitioning the datasets in Benchmark 1 into double the number of clients, with each client having access to only half the bonafide dataset and samples from a single attack (either print or replay). 

\section{Conclusion}
This work addressed the problem of learning generalized FacePAD models based on data from multiple source domains, without pooling all the data into a single location. A novel framework called FedSIS was proposed for this task based on the ViT architecture. An intermediate representation sampling scheme (consisting of a block sampler and adapter network) was introduced to effectively exploit intermediate features of a ViT for FacePAD and serve as a feature augmentation strategy. The resulting ViT-H-IS architecture was trained using a federated split learning strategy. Together, both these contributions enable FedSIS to achieve state-of-the-art domain generalization performance, while preserving privacy. 

\noindent \textbf{Limitations}: The transmission of patch tokens and pseudo-class tokens in each iteration as well as tokenizer and classification head parameters in every unifying round introduces a high communication burden and may result in some privacy leakage. This is a key limitation of the FedSIS framework that needs to be mitigated in the future. Moreover, the current framework does not support continual learning of the model, where clients can join and leave the protocol dynamically. It also needs to be evaluated on unseen attack types not encountered during training. 

{\small
\bibliographystyle{ieee}
\bibliography{egbib}
}

\clearpage
\begin{center}
\textbf{\large Supplementary Material}
\end{center}
Dataset and implementation details are presented in section \ref{sec:dataset_details}. The statistical robustness of the proposed method is analyzed in section \ref{sec:robustness}. Finally, the results of additional ablation studies are reported in section \ref{sec:suppl_ablations}.

\begin{table*}[!t]
\centering \small
\caption{Cross-domain performance of centralized and federated learning methods for Benchmark 1. We run each of our experiments for 3 times under different seeds and report the mean and standard deviation values.}
\label{tab:cross_domain_mcio_suppl}
\setlength{\tabcolsep}{3pt}
      \scalebox{0.55}[0.55]{
	\begin{tabular}{llccccccccccccccccc}
	\multicolumn{17}{c}{} \\ 
        \toprule
	  & \multirow{3}{*}{\bf ~~Method~} & \multicolumn{3}{c}{\textbf{OCI} $\rightarrow$ \textbf{M}}  && \multicolumn{3}{c}{\textbf{OMI} $\rightarrow$ \textbf{C}}  &&  \multicolumn{3}{c}{\textbf{OCM} $\rightarrow$ \textbf{I}} && \multicolumn{3}{c}{\textbf{ICM} $\rightarrow$ \textbf{O}} && \textbf{Avg.} \\
	 \cmidrule{3-5} \cmidrule{7-9} \cmidrule{11-13} \cmidrule{15-17} \cmidrule{18-19} 
	 && \multirow{2}{*}{~HTER~} & \multirow{2}{*}{~AUC~} & TPR@ && \multirow{2}{*}{~HTER~} & \multirow{2}{*}{~AUC~} & TPR@ && \multirow{2}{*}{~HTER~} & \multirow{2}{*}{~AUC~} & TPR@ && \multirow{2}{*}{~HTER~} & \multirow{2}{*}{~AUC~} & TPR@ && \multirow{2}{*}{~HTER~} \\
	 &&&& ~FPR=$1\%$~ &&&& ~FPR=$1\%$~ &&&& ~FPR=$1\%$~ &&&& ~FPR=$1\%$~ \\ 
    \midrule
     & ~~ViT (ECCV' 22) \cite{huang2022adaptive}             & {1.58} & {99.68} & {96.67} && 5.70 & 98.91 & 88.57 && 9.25 & 97.15 & 51.54 && {7.47} & {98.42} & {69.30} && 6.00 \\      \multirow{-2}{*}{Centralized}
     & ~~ViT-H \color{blue}         & 7.81 $\pm$ 1.56 & 97.76 $\pm$ 0.73 & 78.33 $\pm$ 4.71 && 2.64 $\pm$ 0.43 & 99.34 $\pm$ 0.24 & 88.57 $\pm$ 0.59 && 16.08 $\pm$ 1.13 & 90.13 $\pm$ 1.28 & 4.10 $\pm$ 3.22 && 7.13 $\pm$ 0.52 & 97.84 $\pm$ 0.11 & 59.76 $\pm$ 3  && 8.41 $\pm$ 0.91  \\
     & ~~ViT-H + IS       & $2.70 \pm 0.16$ & $99.30 \pm 0.21$  & $83.10 \pm 3.51$  &&  $2.90 \pm 0.01$ & $99.58 \pm 0.07$  & $89.71 \pm 0.96$  && $5.44 \pm 1.01$ &  $99.19 \pm 0.12$ & $74.07 \pm 3.16$ && $11.19 \pm 0.11$  & $95.70 \pm 0.20$  & $40.60 \pm 0.66$ && $5.56 \pm 0.32$  \\
     \midrule
     & ~~FedAVG \cite{FedAvg}         & $8.69 \pm 0.75$ & $97.80 \pm 0.39$ & $77.77 \pm 0.78$ && $1.67 \pm 0.22$ & $99.50 \pm 0.46$ & $95.27 \pm 0.32$ && $20.96 \pm 0.48$ & $84.08 \pm 0.68$ & $3.11 \pm 0.14$  && $9.30 \pm 0.30$  &  $95.60 \pm 0.24$ & $28.03 \pm 0.00$  && $10.15 \pm 0.44$  \\
     & ~~FedProx \cite{Fedprox}       & $7.64 \pm 1.67$ & $97.61 \pm 0.50$ & $71.11 \pm 3.42 $ && $1.98 \pm 0.0$  & $99.75 \pm 0.01$ &  $97.38 \pm 0.34$ && $24.55 \pm 0.10$ & $81.31 \pm 0.13 $  & $1.28 \pm 0.36$& & $17.04 \pm 0.12$ & $91.58 \pm 0.10$ & $31.55 \pm 0.72$ && $12.80 \pm 0.47 $\\
     & ~~MOON \cite{moon} & $7.91 \pm 1.01$  & $97.99 \pm 0.93$  & $75.0 \pm 0.3$ & & $1.98 \pm 0.01$ & $99.818 \pm 0.31$ & $94.28 \pm 0.32 $ & & $21.49 \pm 0.23 $  & $85.19 \pm 0.73 $ & $10.76 \pm 0.91 $ && $11.12 \pm 1.01 $  & $95.10 \pm 0.56 $ & $20.30 \pm 5.54 $ & & $10.62 \pm 0.39 $  \\
     & ~~FeSTA \cite{park2021federated} & $7.81 \pm 1.56$  & $97 \pm 0.77$  & $53.33 \pm 8.50 $ && $7.82 \pm 3.02$ & $97.49 \pm 1.105$ & $65.24 \pm 12.94$ && $9.10 \pm 2.03$  & $96.15 \pm 1.43$ & $40.26 \pm 22.34$ && $17.84 \pm 1.18$  & $89.74 \pm 1.25$ & $15.30 \pm 7.27$ && $10.64 \pm 1.95$  \\
     \rowcolor{yellow!15}
     \multirow{-4}{*}{Federated}
     & ~~FedSIS (Ours)        & $\textbf{1.22} \pm \textbf{0.35}$ & $99.66 \pm 0.10$ & $96.11 \pm 5.50$ && $\textbf{1.39} \pm \textbf{0.00}$ & ${99.67} \pm {0.08}$ & ${96.66} \pm {0.89}$ & & $\textbf{3.36} \pm \textbf{0.29}$ & ${99.18} \pm {0.27}$ & ${75.12} \pm {6.29}$ && $9.52 \pm 0.19$ & $96.25 \pm 0.17$ & $40.23 \pm 3.68$  & & $\textbf{3.91} \pm \textbf{0.11}$ \\
    \bottomrule
	\end{tabular}
}
\end{table*}

\begin{table*}[!t]
\centering \small
\caption{Cross-domain performance of centralized and federated learning methods for Benchmark 2. We run each of our experiments for 3 times under different seeds and report the mean and standard deviations value.}
\label{tab:cross_domain_wcs_supp}
\setlength{\tabcolsep}{3pt}
      \scalebox{0.7}[0.7]{
	\begin{tabular}{llccccccccccccc}
	\multicolumn{13}{c}{} \\ 
        \toprule
	  & \multirow{3}{*}{\bf ~~Method~} & \multicolumn{3}{c}{\textbf{CS} $\rightarrow$ \textbf{W}}  && \multicolumn{3}{c}{\textbf{SW} $\rightarrow$ \textbf{C}}  &&  \multicolumn{3}{c}{\textbf{WC} $\rightarrow$ \textbf{S}} & & \textbf{Avg.} \\
	 \cmidrule{3-5} \cmidrule{7-9} \cmidrule{11-13} 
	 & & \multirow{2}{*}{~HTER~} & \multirow{2}{*}{~AUC~} & TPR@ & & \multirow{2}{*}{~HTER~} & \multirow{2}{*}{~AUC~} & TPR@ && \multirow{2}{*}{~HTER~} & \multirow{2}{*}{~AUC~} & TPR@ && \multirow{2}{*}{~HTER~} \\
	 &&&& ~FPR=$1\%$~ &&&& ~FPR=$1\%$~ &&&&  ~FPR=$1\%$~ \\ 
    \midrule
     & ~~ViT (ECCV' 22) \cite{huang2022adaptive}             & 7.98 & 97.97 & 73.61 &&  11.13& 95.46 & 47.59 && 13.35 &94.13  &  49.97 &&  10.82\\				
     \multirow{-2}{*}{Centralized}
     & ~~ViT-H         & 8.31 $\pm$ 2.46 &  97.71  $\pm$ 1.16& 64.22  $\pm$ 6.36 &&  15.68 $\pm$ 1.22 &  92.99 $\pm$ 1.15 &  48.05 $\pm$ 0.95  && 16.89 $\pm$ 1.22 &  89.97 $\pm$ 0.94 & 32.43 $\pm$ 10.39 &&   13.63 $\pm$ 1.64\\			
     								
     \midrule
     & ~~FedAVG \cite{FedAvg}         & 20.18  $\pm$ 0.93& 89.47 $\pm$ 0.43 & 45.15 $\pm$ 1.15 && 25.94 $\pm$ 0.39 & 82.05 $\pm$  0.44& 22.18 $\pm$ 1.68 && 13.52  $\pm$ 0.29& 93.70 $\pm$ 0.14 & 43.06 $\pm$  13.74 && 19.88 $\pm$ 0.53\\
     								
     & ~~FedProx \cite{Fedprox}       &18.01 $\pm$ 3.32  & 90.81 $\pm$ 2.48 &54.11 $\pm$ 11.79  &&26.98 $\pm$ 0.70 &  81.09 $\pm$ 0.82& 25.95 $\pm$ 4.47 && 12.02 $\pm$ 1.34 & 94.49 $\pm$ 1.33  &  52.04 $\pm$ 9.80 && 14.25 $\pm$ 1.34\\	
     & ~~FeSTA \cite{park2021federated} &  8.49 $\pm$  1.60&  96.79$\pm$ 1.04 & 61.31 $\pm$ 16.26&& 19.15 $\pm$ 3.51& 89.81$\pm$ 3.05 & 32.02$\pm$ 7.06 && 9.60$\pm$ 0.25  & 96.80$\pm$ 0.16 &61.76 $\pm$ 0.33 && 12.41 $\pm$ 1.78\\    	
     
     \rowcolor{yellow!15}
     \multirow{-4}{*}{Federated}
     & ~~FedSIS (Ours)        & \textbf{5.84} $\pm$ \textbf{1.20} &98.81 $\pm$ 0.70 &77.74 $\pm$ 7.50 && \textbf{14.50} $\pm$ 1.03 & 93.80  $\pm$ 1.01&39.89 $\pm$ 5.01  && 11.03 $\pm$ 0.54 & 95.65 $\pm$ 0.32 &55.797  $\pm$ 1.90 &&  \textbf{10.45} $\pm$ \textbf{0.99}\\
     								
    \bottomrule
	\end{tabular}
}
\end{table*}

\section{Dataset and Implementation Details}
\label{sec:dataset_details}

\subsection{Datasets in Benchmark 1} 
We evaluate our method using MSU-MFSD \textbf{(M)} \cite{7031384}, CASIA-MFSD \textbf{(C)} \cite{6199754}, Idiap Replay Attack \textbf{(I)} \cite{6313548}, and OULU-NPU \textbf{(O)} \cite{7961798} datasets. 
\textbf{MSU-MFSD} contains 35 unique subjects and a total of 280 video recordings collected in an indoor scenario using two cameras with different resolutions. The bonafide samples are collected using a laptop and an Android camera. The attack samples contain two qualities of replay attack and one print attack with high-quality photos. The replay attacks are collected using a Canon camera (replayed on an iPad) and an iPhone camera (replayed on itself). The high-quality photos are printed on A3 paper to produce print attacks.
\textbf{CASIA-MFSD} contains 50 unique subjects and a total of 600 video recordings collected under natural scenes. It contains three different image qualities; low, normal, and high. The attack samples include print (warped photo attack, cut photo attack), and replay (replayed on a tablet) attacks, thus creating a total of 9 attack videos per identity. Subsequently, there are 3 real videos per identity, one for each image quality.
\textbf{Idiap Replay-Attack} contains 50 unique subjects and a total of 1300 video recordings. They are collected under two different lighting environments; lamp illuminated with a uniform background and day-light illuminated with a complex scene. The print and replay attacks are generated similarly to that of MSU-MFSD with different capture devices. They additionally provide two new environment variations, where these attack materials are presented either on fixed support or by hand. 
\textbf{OULU-NPU} contains 55 unique subjects and a total of 3600 videos captured in three sessions, each with different illumination conditions and background scenes. The bonafide samples are recorded using the front camera of six different mobile phones. The attack samples contain print and replay attacks generated using two printers and two video players, thus creating a more diverse and larger set of samples than the other datasets.

\subsection{Datasets in Benchmark 2} 
We evaluate our method using WMCA \textbf{(W)} \cite{8714076}, CASIA-CeFA \textbf{(C)} \cite{9423056, liu2021cross}, and CASIA-SURF \textbf{(S)} \cite{8995504, zhang2019dataset} datasets. 
\textbf{WMCA} contains 72 unique subjects and a total of 1941 short video recordings. The dataset provides a wide range of attacks, namely; print, replay (Tablet), partial (Glasses), and mask (Plastic, Silicone, Paper, and Mannequin). It also provides multiple modalities/channels (RGB, Depth, Infrared and, Thermal).
\textbf{CASIA-CeFA} contains 1607 unique subjects covering 3 ethnicities (Africa, East Asia, and Central Asia) and a total of 23538 videos. The dataset provides 3 modalities/channels (RGB, Depth, and, Infrared) with 4 attack types (print, replay, and mask; 3D print, silica gel) under indoor and outdoor lighting conditions.
\textbf{CASIA-SURF} contains 1000 unique subjects and a total of 21000 video recordings. The dataset provides 3 modalities/channels (RGB, Depth, and Infrared) for each sample. It contains 6 different types of print attack. The background regions are completely removed, and the print attacks have randomly cut eyes, nose, or mouth areas.
 However, following \cite{huang2022adaptive}, we use only the print and replay attack in the RGB channel for each of these three datasets.

\begin{table*}[!t]
\centering \small
\caption{Cross-domain performance when training with one attack type per client. CelebA-Spoof = \{Included, Excluded\} indicates whether CelebA-Spoof is used as the auxiliary dataset. \emph{Combined} : Each client has the same attack types (both print and replay), but with different environmental factors. \emph{Split} : Each client has a single attack type (either print or replay) and different environmental factors.  We run each of our experiments for 3 times with different seeds and report the mean values.}
\label{tab:client_wise_attack_mcio}
\setlength{\tabcolsep}{3pt}
      \scalebox{0.80}[0.80]{
	\begin{tabular}{ccccccccccccccccccc}
	\multicolumn{16}{c}{} \\ 
        \toprule
	  \multirow{3}{*}{\bf ~~CelebA-Spoof~ }& \multirow{3}{*}{\bf ~~Attacks~} & \multicolumn{3}{c}{\textbf{OCI} $\rightarrow$ \textbf{M}}  && \multicolumn{3}{c}{\textbf{OMI} $\rightarrow$ \textbf{C}}  &&  \multicolumn{3}{c}{\textbf{OCM} $\rightarrow$ \textbf{I}} && \multicolumn{3}{c}{\textbf{ICM} $\rightarrow$ \textbf{O}} && \textbf{Avg.} \\
	 \cmidrule{3-5} \cmidrule{7-9} \cmidrule{11-13} \cmidrule{15-17} 
	 && \multirow{2}{*}{~HTER~} & \multirow{2}{*}{~AUC~} & TPR@ && \multirow{2}{*}{~HTER~} & \multirow{2}{*}{~AUC~} & TPR@ && \multirow{2}{*}{~HTER~} & \multirow{2}{*}{~AUC~} & TPR@ && \multirow{2}{*}{~HTER~} & \multirow{2}{*}{~AUC~} & TPR@ && \multirow{2}{*}{~HTER~} \\
	 &&&& ~FPR=$1\%$~ &&&& ~FPR=$1\%$~ &&&& ~FPR=$1\%$~ &&&& ~FPR=$1\%$~ \\ 
    \midrule
     &~~\emph{Combined}         & ${1.22} $ & ${99.66}$ & ${96.11}$ & & ${1.39}$ & ${99.67}$ & ${96.66} $ & & ${3.36}$ & ${99.18} $ & ${75.12} $ && $9.52$ & ${96.25}$ & ${40.23}$  & & ${3.91}$ \\
     \multirow{-2}{*}{Included}
     &~~\emph{Split}        & 1.13 & 99.82 & 99.96 && 1.35 & 99.78 & 96.99 && 4.65 & 98.93 & 78.21 && 11.74 & 94.18 & 26.94 && 4.74 \\
     \midrule
     &~~\emph{Combined}     & {2.25} & 99.42 & 72.16 && {5.75} & 98.49 & 70.77 && 7.39 & 97.44 & 52.52 && 10.68 & 95.80 & 45.57 && {6.43}  \\
     \multirow{-2}{*}{Excluded} 
     &~~\emph{Split}        & 2.44 & 99.59 & 89.44 && 5.89 & 98.49 & 75.71 && 6.91 & 98.17 & 65.63 && 12.97 & 94.31 & 35.72 && 6.94 \\
    \bottomrule
	\end{tabular}
}
\end{table*}

\subsection{Frame Sampling Strategy}
We follow the frame sampling strategy and train-test split proposed in \cite{huang2022adaptive} for benchmarks 1 and 2.

\noindent \textbf{Benchmark 1}:  We sample one frame from each video to train the model and evaluate using two frames in each video. We report the values of the evaluation metrics by averaging the probabilities across frames for each video. 

\noindent \textbf{Benchmark 2}: For WMCA and CASIA-CeFA, we sample ten frames equidistantly from each video and also evaluate the model using ten frames. For CASIA-SURF, we use all the frames from the dataset.

\section{Robustness Across Multiple Runs}
\label{sec:robustness}
As highlighted in \cite{huang2022adaptive}, generalization performance on the unseen target domain can fluctuate among different checkpoints for two main reasons. Firstly, training large models with few samples leads to catastrophic forgetting causing training instability. Secondly, the large domain gap between the source and target domains can cause high uncertainty as the target samples can lie close to the decision boundary. Furthermore, the non-deterministic (due to random block sampling) nature of predictions in the proposed FedSIS method may also lead to fluctuations. To conduct a more reliable evaluation that accounts for these uncertainties, we run each of our experiments with three different seeds and report the mean and standard deviation values.
As can be seen from Tables \ref{tab:cross_domain_mcio_suppl} and \ref{tab:cross_domain_wcs_supp}, the average standard deviation values for both the benchmarks are very low (0.11\% for benchmark 1 and 0.99\% for benchmark 2) indicating that the proposed approach is robust to the instabilities during training. 

\begin{figure*}[t!]
   \centering{   \includegraphics[width=\textwidth]{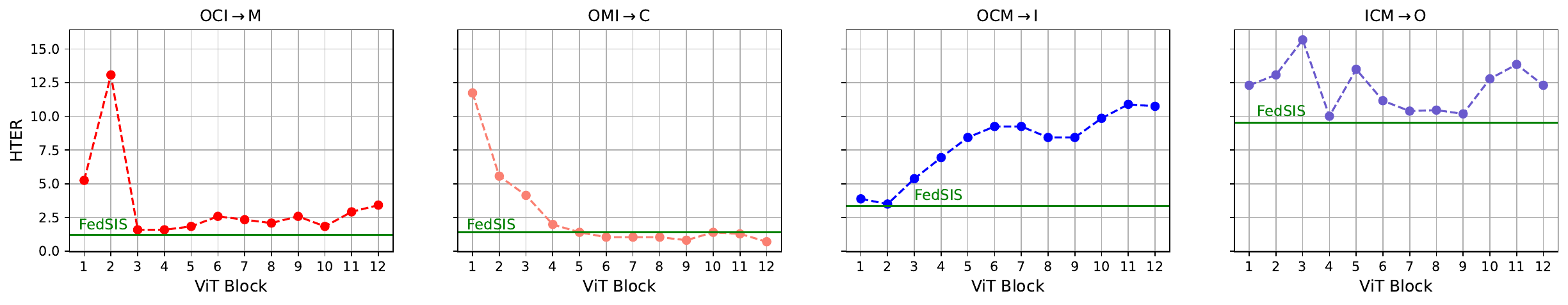}
   }
   \caption{Cross-domain HTER(\%) when block sampler is removed and intermediate representations are derived from a single pre-determined block.} 
   \label{fig:separate_blocks}
\end{figure*}

\begin{figure*}[t]
  \centering
  \begin{subfigure}[b]{0.47\linewidth}
    \centering\includegraphics[width=0.8\linewidth]{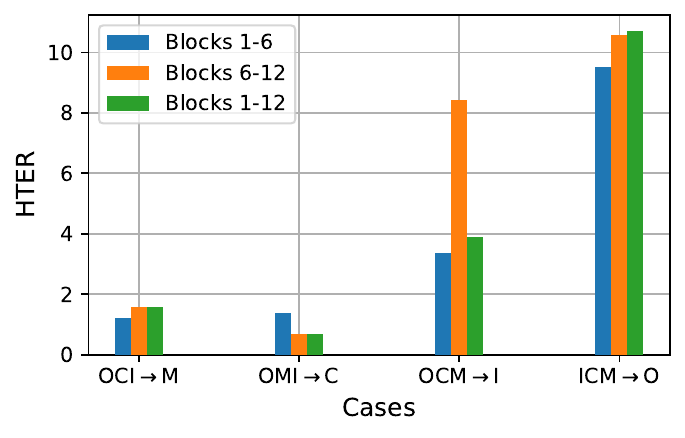}
    \label{fig:sub1}
  \end{subfigure}
  \begin{subfigure}[b]{0.47\linewidth}
    \centering\includegraphics[width=0.8\linewidth]{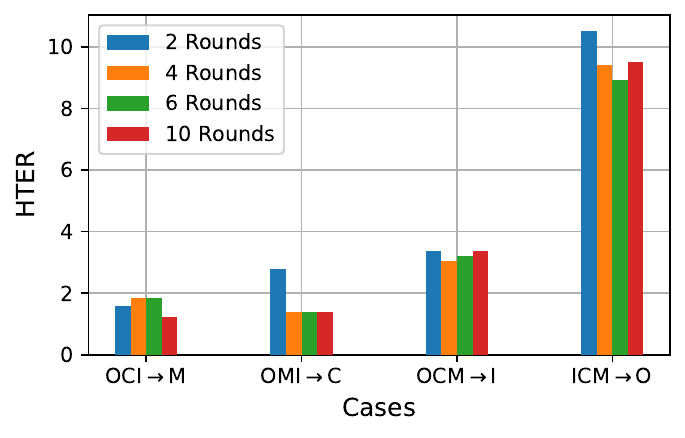}
    \label{fig:sub2}
  \end{subfigure}
  \caption{\textbf{Left}: FedSIS performance for different ranges of ViT sampling blocks. \textbf{Right}: FedSIS performance with various unifying rounds for the federation of the clients' heads and tails.}\label{fig:ablation_blocks_unifyingRound}
\end{figure*}

\begin{table*}[!t]
\centering \small
\caption{Comparing the proposed approach with more centralized learning methods from the literature. FedSIS$^{*}$ indicates not using CelebA-Spoof both at the pre-training and collaborative training stages, while FedSIS indicates using at both these stages. For FedSIS and FedSIS$^{*}$, we run the experiment 3 times and report the mean performance.
}
\label{tab:suppl_fedsis_vs_central}
\setlength{\tabcolsep}{3pt}
      \scalebox{0.8}[0.80]{
	\begin{tabular}{llccccccccccccccccc}
	\multicolumn{16}{c}{} \\ 
        \toprule
	  & \multirow{3}{*}{\bf ~~Method~} & \multicolumn{3}{c}{\textbf{OCI} $\rightarrow$ \textbf{M}}  && \multicolumn{3}{c}{\textbf{OMI} $\rightarrow$ \textbf{C}}  &&  \multicolumn{3}{c}{\textbf{OCM} $\rightarrow$ \textbf{I}} && \multicolumn{3}{c}{\textbf{ICM} $\rightarrow$ \textbf{O}} && \textbf{Avg.} \\
	 \cmidrule{3-5} \cmidrule{7-9} \cmidrule{11-13} \cmidrule{15-17} 
	 && \multirow{2}{*}{~HTER~} & \multirow{2}{*}{~AUC~} & TPR@ && \multirow{2}{*}{~HTER~} & \multirow{2}{*}{~AUC~} & TPR@ && \multirow{2}{*}{~HTER~} & \multirow{2}{*}{~AUC~} & TPR@ && \multirow{2}{*}{~HTER~} & \multirow{2}{*}{~AUC~} & TPR@ && \multirow{2}{*}{~HTER~} \\
	 &&&& ~FPR=$1\%$~ &&&& ~FPR=$1\%$~ &&&& ~FPR=$1\%$~ &&&& ~FPR=$1\%$~ \\ 
    \midrule
    
     &~~SSAN-R (CVPR' 22) ~\cite{wang2022domain}       & 6.67  & 98.75 & -- && 10.00 & 96.67 & -- && 8.88  & 96.79 & -- && 13.72 & 93.63 & -- && 9.80\\
     &~~PatchNet (CVPR' 22) ~\cite{wang2022patchnet}       &  7.10 & 98.46 & -- && 11.33 & 94.58 & -- && 13.40 & 95.67 & -- && 11.82 & 95.07 & -- && 10.90\\
     &~~GDA (ECCV' 22) ~\cite{zhou2022generative}       & 9.20  & 98.00 & -- && 12.20 & 93.00 & -- && 10.00 & 96.00 & -- && 14.40 & 92.60 & -- && 11.45\\
     &~~TransFAS (TBIOM' 22) ~\cite{wang2022face}       &  7.08 & 96.69 & -- && 9.81 & 96.13 & -- && 10.12 & 95.53 & -- && 15.52 & 91.10 & -- && 10.63\\
     &~~DiVT-M (WACV' 23) \cite{liao2023domain}       & 2.86  & 99.14 & -- && 8.67  & 96.62 & -- && {3.71}  & 99.29 & -- && 13.06 & 94.04 & -- && 7.07\\
     &~~SA-FAS (CVPR' 23) \cite{Sun2023Mar}       & 5.95 & 96.55 & - && 8.78 & 95.37 & - && {6.58} & 97.54 & - && 10.00 & 96.23 & - && 7.81 \\
     &~~IADG  (CVPR' 23) \cite{zhou2023instance}     & 5.41 & 98.19 &  - && 8.70 & 96.44 & - && 10.62 & 94.50 & - && {8.86} & 97.14 & - && 8.39 \\
     \midrule
     & ~~FedPAD (TNNLS' 22) \cite{shao2022federated} & 19.45 & 90.24  & - && 42.27 & 70.49 & - && 32.53  & 73.58 & - &&  34.44 & 71.74 & - && 32.17   \\
     & ~~FedGPAD (TNNLS' 22) \cite{shao2022federated} & 12.73 & 91.25  & - && 28.69 & 80.58 & - && 10.97 & 95.34 & - && 21.95 & 89.85 & -  && 18.59  \\
     \midrule
    \rowcolor{yellow!15}
     &~~FedSIS$^{*}$  (Ours)        & {2.25} & 99.42 & 72.16 && {5.75} & 98.49 & 70.77 && 7.39 & 97.44 & 52.52 && 10.68 & 95.80 & 45.57 && {6.43}  \\
    \rowcolor{yellow!15}
     & ~~FedSIS (Ours)        & ${1.22} $ & ${99.66}$ & ${96.11}$ & & ${1.39}$ & ${99.67}$ & ${96.66} $ & & ${3.36}$ & ${99.18} $ & ${75.12} $ && $9.52$ & ${96.25}$ & ${40.23}$  & & ${3.91}$ \\
    
    \bottomrule
	\end{tabular}
}
\end{table*}

\section{Ablation Studies}
\label{sec:suppl_ablations}

\noindent \textbf{Impact of sampling}: To understand the role of the block sampler, we conducted an experiment that excludes the sampler and takes the output from only a specific block of the ViT feature encoder. Fig. \ref{fig:separate_blocks} indicates that the choice of block significantly impacts the performance, with different blocks performing well for different scenarios. For example, we observed that the later blocks exhibit improved performance in the OCI $\rightarrow$ M scenario, whereas initial blocks yield better performance in the OCM $\rightarrow$ I scenario. The results also indicate that the performance of individual blocks, without the block sampler, was either worse or comparable to that of FedSIS as shown in Fig \ref{fig:separate_blocks}. These findings emphasize the critical role of intermediate features in enhancing FacePAD performance, which conforms with the observations in \cite{rouqiah_paper, wang2022face}.

\noindent \textbf{Block Sampling Choice}:
Fig. \ref{fig:ablation_blocks_unifyingRound} (\textbf{Left}) shows the effect of sampling from only a subset of ViT blocks. We studied the performance of FedSIS with the sampler having access to different sets of ViT blocks from which it could sample. Except for OMI $\rightarrow$ C, Fig. \ref{fig:ablation_blocks_unifyingRound} (\textbf{Left}) illustrates that FedSIS achieves higher performance when sampling from blocks $1-6$ of the ViT, instead of $1-12$ or $6-12$. Moreover, comparing the results with the performance of each ViT block in Fig. \ref{fig:separate_blocks} indicates that sampling from blocks ($1-6$) proves the efficacy of the sampler in having a better discriminative model. Finally, opting for the first $6$ blocks also reduces the computational complexity while still achieving better performance than choosing all the ViT blocks.

\noindent \textbf{Unifying Rounds.} 
The unifying rounds $(r_{uni})$ refer to the number of times we perform federation for the tokenizer and classification heads via FedAvg. To assess the effect of federation frequency on the final performance, we conduct several experiments, each with varying federation rounds. We increment the frequency by two and visualize the results for Benchmark 1 in Figure \ref{fig:ablation_blocks_unifyingRound} (\textbf{Right}). Our results indicate that comparable performance can be achieved with different frequencies, with the best performance achieved when the federation occurs every 10 rounds. Utilizing a lower frequency of federation can reduce communication costs between the client and server, providing an additional incentive to select more rounds for the federation.

\noindent \textbf{Effect of One Attack Per Client.}
In Benchmarks 1 and 2, we presented the results of training with heterogeneous (non-iid) data distribution. However, despite the distribution shifts, the types of attack in each of these clients remain the same (print and replay). 
To further challenge the FedSIS framework, we investigate the effect of having only one attack type per client (results shown in Table \ref{tab:client_wise_attack_mcio}). 
This creates a more heterogeneous setting where the heterogeneity comes from both the distribution shift and the attack type. 
Specifically, we split each client into two sub-clients, where each sub-client contains only a non-overlapping subset of bonafide and attack samples. For example, in Benchmark 1 we split MSU-MFSD into two clients, where one contains the print attack samples and half of the bonafide samples, and the other contains the replay attack samples with the remaining bonafide samples. This creates 6 training clients when CelebA-spoof is excluded and 7 when it's included. We observe that even for this extreme heterogeneous setting (denoted as \emph{Split}), our method's performance is on-par with attack-invariant data (denoted as \emph{Combined}). When compared to Table \textcolor{red}{1} in the main paper, we also note that FedSIS trained with attack-variant data outperforms other collaborative learning methods by large margins. These observations indicate that FedSIS could be effective even when each client has different attack types and varying environmental factors.

\end{document}